%% file: iclr2025_conference.tex
\documentclass{article} 
\usepackage{iclr2025_conference,times}

\input{math_commands.tex}

\usepackage{url}
\usepackage{graphicx}
\usepackage{amsmath}
\usepackage{amssymb}
\usepackage{booktabs}
\usepackage{multirow}
\usepackage{tabularx}
\usepackage{makecell}
\usepackage{placeins}
\usepackage{dirtree}
\usepackage{subcaption}
\usepackage{todonotes}
\usepackage{wrapfig}

\usepackage[pagebackref,breaklinks]{hyperref}

\usepackage[capitalize]{cleveref}
\crefname{section}{Sec.}{Secs.}
\Crefname{section}{Section}{Sections}
\Crefname{table}{Table}{Tables}
\crefname{table}{Tab.}{Tabs.}

\usepackage{xcolor}         

\title{A Model Zoo of Vision Transformers}

\author{Damian Falk, Léo Meynent, Florence Pfammatter, Konstantin Schürholt, Damian Borth  \\
  AIML Lab, \\
  School of Computer Science,\\
  University of St.Gallen\\
\texttt{\{first.last\}@unisg.ch} \\
}

\iclrfinalcopy 
\begin{document}

\maketitle

\begin{abstract}
   The availability of large, structured populations of neural networks --- called ``model zoos'' --- has led to the development of a multitude of downstream tasks ranging from model analysis, to representation learning on model weights or generative modeling of neural network parameters. 
   However, existing model zoos are limited in size and architecture and neglect the transformer, which is among the currently most successful neural network architectures.
   We address this gap by introducing the first model zoo of vision transformers (ViT). To better represent recent training approaches, we develop a new blueprint for model zoo generation that encompasses both pre-training and fine-tuning steps, and publish 250 unique models. They are carefully generated with a large span of generating factors, and their diversity is validated using a thorough choice of weight-space and behavioral metrics.
   To further motivate the utility of our proposed dataset, we suggest multiple possible applications grounded in both extensive exploratory experiments and a number of examples from the existing literature. By extending previous lines of similar work, our model zoo allows researchers to push their model population-based methods from the small model regime to state-of-the-art architectures. We make our model zoo available at \texttt{\href{https://github.com/ModelZoos/ViTModelZoo}{github.com/ModelZoos/ViTModelZoo}}.
\end{abstract}

\section{Introduction}
\label{sec:intro}

Recent years have seen growing interest in studying populations of neural networks -- referred to as ``model zoos'' --  to predict properties such as accuracy or hyperparameters used to train the model \citep{unterthinerPredictingNeuralNetwork2020, eilertsenClassifyingClassifierDissecting2020, schurholtModelZoosDataset2022}. 
There are multiple applications to the study of machine learning models from a population perspective: from analyzing the training procedure~\citep{jaderbergPopulationBasedTraining2017,chenLearningUniversalHyperparameter2022,zhou2024MD}, to predicting model properties~\citep{eilertsenClassifyingClassifierDissecting2020, unterthinerPredictingNeuralNetwork2020, kahana2024deep}, learning representations of neural network models~\citep{schurholtHyperRepresentationsPreTrainingTransfer2022, schurholt2024SANE}, generating new weights for models~\citep{knyazevParameterPredictionUnseen2021, schurholtHyperRepresentationsGenerativeModels2022, peeblesLearningLearnGenerative2022}, or averaging models for better generalization or robustness~\citep{izmailovAveragingWeightsLeads2019a, wortsmanModelSoupsAveraging2022, rameModelRatatouilleRecycling2023}.

Previous datasets curated to cover this research area exist, but contain either small CNN models only~\citep{unterthinerPredictingNeuralNetwork2020, schurholtModelZoosDataset2022, dong2022zood, navonEquivariantArchitecturesLearning2023}, or are tailored to very specific applications~\citep{liu2018knowledge, reed2022self, honeggerSparsifiedModelZoo2023, honeggerEurosatModelZoo2023, croce2020robustbench}. While such model zoos do provide a foundation for the research stated above, they miss to represent state-of-the-art model architectures, such as the transformer~\citep{vaswaniAttentionAllYou2017}. This architecture has become state-of-the-art in computer vision~\citep{dosovitskiyImageWorth16x162020}, language modeling~\citep{devlinBERTPretrainingDeep2018, radfordImprovingLanguageUnderstanding2018, touvronLlamaOpenFoundation} and other domains~\citep{radfordRobustSpeechRecognition2022}.

Furthermore, recent training procedures in computer vision applications are usually composed of two parts, starting with a general pre-training stage followed by a more task-specific fine-tuning stage. It is common practice to re-use public pre-trained models and fine-tune them for more specific tasks. Existing model zoos do not represent this kind of training scheme. Thanks to the Hugging Face ``\textit{transformers}'' library and the corresponding model repository~\citep{wolf2020transformers}, large collections of pre-trained and fine-tuned transformer models are nowadays freely available online. Unfortunately, leveraging such an unstructured population remains complex: documentation for these model is often scarce, training is non-standardized and may follow different protocols.

\paragraph{Contributions} Our model zoo directly addresses these gaps, presenting a structured population of ViT models and a large panel of potential applications. To achieve this, we extend the blueprint of previous published works on the topic~\citep{schurholtModelZoosDataset2022}: transformers are usually trained in two steps, first a self-supervised or supervised pre-training step, followed by a supervised fine-tuning step. 
Adapted to this new training protocol, our work presents a computer vision model zoo containing $250$ ViT-S~\citep{touvron2021vit-s} models ($10$ pre-trained and $240$ fine-tuned) with multiple model states collected throughout the training process. Our dataset is presented in a structured format easily accessible to the research community. We include an in-depth diversity analysis of our model zoo that covers both structural/weight space and behavioral metrics. We additionally suggest potential applications for our model zoo, and explore in particular model lineage prediction and model weights averaging, showcasing the potential of this dataset as a challenging testbed for research in the domain of weight space learning.


\section{Model Zoo Generation}
\label{sec:zoo_generation}

\begin{figure}[h]
    \centering
    \includegraphics[width=\columnwidth]{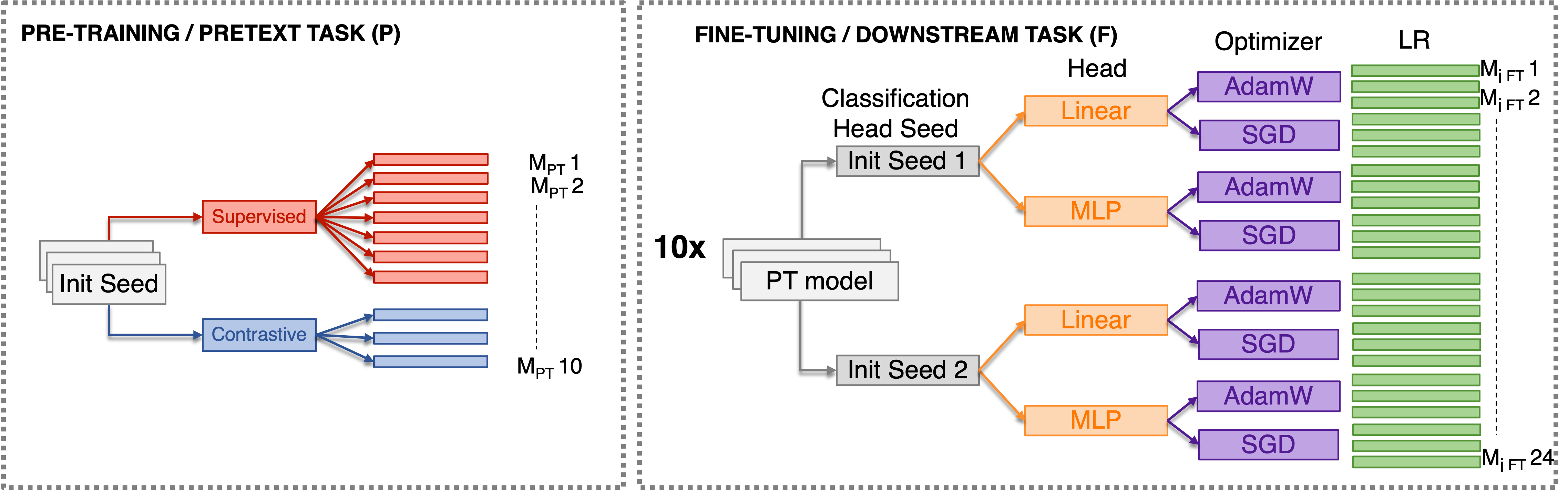}
    \caption{Overview of the model zoo generating factors. On the left, we show the two different training tasks, and multiple pre-training seeds. On the right, we show how each of the pre-trained models is fine-tuned using different configurations forming a hyperparameter grid.}
    \label{fig:trainig_illustration}
\end{figure}

Previous work on model zoos has formalized the generation process of model zoos to make zoos reproducible, and control the variation in hyperparameters as well as the diversity of the models in the zoo~\citep{unterthinerPredictingNeuralNetwork2020, schurholtModelZoosDataset2022}. However, those works mostly focused on small-scale models and convolutional architectures like LeNets and ResNets. In recent years, training recipes have evolved from those used in previous model zoos. For instance, training can be done in two separate steps: (a) pre-training and (b) fine-tuning. In the following, we adapt the model zoo generation scheme to represent this new configuration space of latent generating factors.

To represent the two-stages training procedure, we define two tuples $\{\mathcal{A}_{P}, \mathcal{D}_{P}, \lambda_{P}\}$ and $\{\mathcal{A}_{F}, \mathcal{D}_{F}, \lambda_{F}\}$ that refer to the architecture, the dataset and the set of hyperparameters for pre-training and fine-tuning, respectively. 
While the datasets $\mathcal{D}_P, \mathcal{D}_F$ and architectures $\mathcal{A}_P$ are fixed for a model zoo, we can vary $\mathcal{A}_F$, $\lambda_{P}$ and $\lambda_F$ to define the hyperparameters grid.
\footnote{$\mathcal{A}_F$ and  $\mathcal{A}_P$ are distinct since the classification head architecture needs to be adjusted for the differing number of classes depending on the dataset used. Besides its output dimension, $\mathcal{A}_F$ can also vary in nature, as we use either a linear probe or a Multi-Layer Perceptron (MLP).}

Varying $\lambda_{P}$ allows the comparison of multiple pre-training tasks and the evaluation which form the best basis for fine-tuning. Different pre-training tasks are expected to have a significant impact on the models and create distinct modes.
On the other hand, varying $\mathcal{A}_F$ and $\lambda_{F}$ allows us to evaluate the diversity of models fine-tuned from a same --- or a differing --- pre-trained model. Variations in the different components in $\mathcal{A}_F$ and $\lambda_{F}$ are expected to have varying impact on the diversity; different optimizers may lead to disjunct modes, varying the learning rate to smoother transitions. To model an interesting and representative model population, we choose to vary both types of generating factors.\looseness-1

\begin{table*}[t]
\centering
\small
\caption{Analysis of the performance and diversity of the pretrained models. The $L^2$ distance is calculated between fine-tuned models and their pre-trained ancestor, while the between cluster distance is calculated pairwise between the pretrained models per pre-training task and overall. \looseness-1}
\label{tab:diversity-pt}
\begin{tabular}{ccccccc}
\toprule
Task          & Performance    & \multicolumn{3}{c}{Weight Statistics}                                                                           & \multicolumn{2}{c}{$L^2$ Distance}                                          \\ 
\cmidrule(rl){1-1} \cmidrule(rl){2-2} \cmidrule(rl){3-5} \cmidrule(rl){6-7} 
$\lambda_P$      & Accuracy       & Skewness                          & Kurtosis                          & $L^2$ Norm                                 & \multicolumn{1}{c}{Within Cluster} & \multicolumn{1}{c}{Between Cluster} \\ 
\cmidrule(rl){1-1} \cmidrule(rl){2-2} \cmidrule(rl){3-5} \cmidrule(rl){6-7} 
Supervised  & 68.39$\pm$4.85 & 0.01$\pm$0.02                     & 0.85$\pm$0.24                     & 1636$\pm$248                      & 277$\pm$324                  & 2689$\pm$425                  \\
Contrastive & -              & 0.01$\pm$0.01                     & 3.06$\pm$0.22                     & 2568$\pm$224                      & 501$\pm$346                  & 3640$\pm$164                  \\ 
\cmidrule(rl){1-1} \cmidrule(rl){2-2} \cmidrule(rl){3-5} \cmidrule(rl){6-7} 
\textbf{All}     & -              & \multicolumn{1}{l}{0.01$\pm$0.01} & \multicolumn{1}{l}{1.51$\pm$1.09} & \multicolumn{1}{l}{1916$\pm$505} & 325$\pm$340                  & 2753$\pm$476                   \\ \bottomrule
\end{tabular}
\end{table*}    

\newpage
\paragraph{Generating factors} Our vision transformer model zoo is built with $\mathcal{A}_P =$ {ViT-S}~\citep{touvron2021vit-s}, pre-trained on $\mathcal{D}_P =$ {ImageNet-1k}~\citep{imagenet15russakovsky} and fine-tuned on $\mathcal{D}_F =$ {CIFAR-100}~\citep{cifar}. We adopt fixed pre-training hyperparameters, $\lambda_P$, from prior research \citep{beyer2022better, touvron2021vit-s, steinerHowTrainYour2021}, balancing both strong performance and diversity within the model zoo. To have a more varied set of pre-trained models and enable distinct modes within the zoo, we vary $\lambda_P$ by either pre-training supervised on {ImageNet-1k}, or self-supervised using the SimCLR~\citep{chen2020simclr} contrastive learning framework. The training scheme is illustrated in Figure~\ref{fig:trainig_illustration}. We aim for 10 different pre-trained models, and since supervised pre-training is more common, we use 7 fixed seeds for the supervised task, and 3 for the self-supervised contrastive task.
After pre-training, we fine-tune each backbone with systematically varied hyperparameters $\lambda_F$, resulting in 24 fine-tuned variations per pre-trained model. Although not every combination of $\lambda_F$ is expected to achieve competitive performance, we include all variations to enable systematic studies on the effects of altering single parameters. We summarize pre-training configurations and fine-tuning variations in Appendix~\ref{sec:generatingfactors_vit}.
%
%
%

\section{Model Zoo Analysis}
\label{sec:zoo_analysis}
\label{sec:performance}
In this Section, we analyze the performance and behavioral diversity of the generated model zoos. To that end, we largely follow a model diversity analysis similar to previous model zoos \citep{schurholtModelZoosDataset2022}. Although our zoo contains multiple training epochs for each model, our analysis focuses on the last one. To measure models' generalization, we calculate the generalization gap $\text{GGap} = \text{Acc}_{\text{Train}} - \text{Acc}_{\text{Test}}$. We also measure the behavioral diversity of the models via agreement. We report the performance and diversity metrics of the models on the pre-training task ImageNet-1k in Table \ref{tab:diversity-pt}, and on the fine-tuning task CIFAR-100 in Table \ref{tab:diversity-refined}.

\begin{table*}[t]
\centering
\small
\caption{Analysis of the performance and diversity of the refined (validation accuracy above 65\%) ViT-S model zoo evaluated on CIFAR-100. Mean$\pm$std and generalization gap values reported in percent per configuration after the last epoch of training. Agreement is computed pairwise on the validation set per configuration and overall, values reported in \%. The higher the agreement, the lower the diversity. Weight statistics are computed per model and averaged per isolated configuration. The $L^2$ distance is calculated between the finetuned models and their ancestors. Metrics for the raw zoo including all models are included in the Appendix~\ref{sec:ft-acc-overview} Table \ref{tab:pt-acc-ft}. \looseness-1}

\setlength{\tabcolsep}{2pt}\label{tab:diversity-refined}

\begin{tabularx}{\linewidth}{ccccccccc}
\toprule

\multicolumn{2}{c}{Generating Factors} & \multicolumn{2}{c}{Performance} & Agreement                  & \multicolumn{4}{c}{Weight Statistics}                                    \\ 
\cmidrule(rl){1-2} \cmidrule(rl){3-4} \cmidrule(rl){5-5} \cmidrule(rl){6-9} 
$\lambda_F$           & Config      & Accuracy       & GGap           & \( \kappa_{\text{aggr}} \) & Skewness        & Kurtosis      & $L^2$ Norm         & $L^2$ Distance    \\ 
\cmidrule(rl){1-2} \cmidrule(rl){3-4} \cmidrule(rl){5-5} \cmidrule(rl){6-9} 
\multirow{2}{*}{Pre-tr.} & Sup.  & 78.88$\pm$5.30 & 8.99$\pm$7.58  & 78.23$\pm$4.92             & 0.011$\pm$0.015 & 0.82$\pm$0.23 & 1689$\pm$231 & 277$\pm$324 \\
                         & Contr. & 74.83$\pm$4.96 & 11.85$\pm$6.39 & 75.59$\pm$4.50             & 0.008$\pm$0.008 & 2.99$\pm$0.19 & 2613$\pm$192 & 501$\pm$346 \\ 
\cmidrule(rl){1-2} \cmidrule(rl){3-4} \cmidrule(rl){5-5} \cmidrule(rl){6-9} 
\multirow{2}{*}{Head}    & MLP         & 78.36$\pm$5.35 & 10.12$\pm$6.77 & 77.29$\pm$4.73             & 0.009$\pm$0.014 & 1.32$\pm$0.95 & 1904$\pm$457 & 342$\pm$341 \\
                         & Linear      & 77.72$\pm$5.59 & 9.13$\pm$7.54  & 75.90$\pm$5.35             & 0.009$\pm$0.014 & 1.25$\pm$0.89 & 1868$\pm$434 & 309$\pm$341 \\ 
\cmidrule(rl){1-2} \cmidrule(rl){3-4} \cmidrule(rl){5-5} \cmidrule(rl){6-9} 
\multirow{2}{*}{Optim.}  & AdamW       & 80.06$\pm$4.59 & 13.55$\pm$3.96 & 78.29$\pm$5.08             & 0.009$\pm$0.013 & 1.46$\pm$1.02 & 1986$\pm$466 & 448$\pm$324 \\
                         & SGD         & 72.76$\pm$3.88 & -0.61$\pm$2.45 & 78.41$\pm$4.05             & 0.008$\pm$0.015 & 0.83$\pm$0.22 & 1624$\pm$225 & 3.20$\pm$1.02     \\ 
\cmidrule(rl){1-2} \cmidrule(rl){3-4} \cmidrule(rl){5-5} \cmidrule(rl){6-9} 
\multirow{3}{*}{LR}      & 3E-3        & 77.78$\pm$2.29 & 11.07$\pm$9.20 & 76.54$\pm$3.29             & 0.009$\pm$0.014 & 1.18$\pm$0.84 & 1922$\pm$433 & 513$\pm$430 \\
                         & 1E-3        & 77.75$\pm$6.77 & 10.18$\pm$8.33 & 76.52$\pm$5.99             & 0.009$\pm$0.014 & 1.28$\pm$0.93 & 1838$\pm$443 & 262$\pm$178 \\
                         & 1E-4        & 78.86$\pm$7.01 & 6.18$\pm$2.98  & 78.33$\pm$7.90             & 0.009$\pm$0.014 & 1.47$\pm$1.02 & 1893$\pm$468 & 89.9$\pm$9.40    \\ 
\cmidrule(rl){1-2} \cmidrule(rl){3-4} \cmidrule(rl){5-5} \cmidrule(rl){6-9} 
\textbf{All}             & -           & 78.02$\pm$5.49 & 9.60$\pm$7.18  & 76.50$\pm$5.18             & 0.009$\pm$0.014 & 1.28$\pm$0.92 & 1885$\pm$444 & 325$\pm$340 \\ 

\bottomrule
\end{tabularx}
\vspace{-8pt}
\end{table*}

\subsection{Performance}
\label{sec:performance_det}
The results show high performance of our models on both pre-training and fine-tuning task, which validates the training strategies. Although the population is optimized for diversity rather than performance, the models reach up to $72.4\%$ validation accuracy on the pre-training dataset Imagenet-1k. On the fine-tuning dataset CIFAR-100, our models achieve up to $85.1\%$ validation accuracy. 
The performance of the well trained models in the zoo is in line with related work on training similar ViT models:
$72.3\%$ on CIFAR-100 (with ViT-T)~\citep{xuInitializingModelsLarger2023a};
$74.4\%$ on ImageNet-1k and $85.7\%$ on CIFAR-100 (with ViT-S/16)~\citep{chenWhenVisionTransformers2021}; $66.8\%$ on ImageNet-1k with the original training-setup, $71.6\%$ when introducing additional augmentations up to $76.5\%$ by further  modifications to the architecture $\mathcal{A_P}$ (with ViT-S/16) \citep{beyer2022better}; 
$77.5\%$ on ImageNet-1k and $86.9\%$ on CIFAR-100 (with ViT-S/16, and 300 pre-training epochs compared to our 90)~\citep{steinerHowTrainYour2021}.
Naturally, the variations in the training strategy we introduced have an impact on model performance, and not all of the models on our hyperparameters grid perform as well as the better ones described above. We include performance and structural (i.e. weight-space) metrics of the influence of our training hyperparameters in Tables~\ref{tab:diversity-pt} and~\ref{tab:diversity-refined}. 
%

\newpage
\subsection{Diversity}
\label{sec:diversity}

The design space of our model zoos is chosen to create diverse models, rather than only high-performing ones. In this Section, we evaluate behavioral diversity by computing the distribution of performance overall and within specific configurations, as well as model agreement. We supplement these data-driven metrics with model weight statistics: skewness, kurtosis, as well as the $L^2$ norm in Table \ref{tab:diversity-refined}. To corroborate the notion of distinct modes in weight space as a result of the generation scheme, we also compute pair-wise $L^2$ distances between and within clusters in Table~\ref{tab:diversity-pt}. Specifically, we compute the $L^2$ distance between the pre-trained models' clusters, as well as between the pre-trained models and their fine-tuned derivatives.\looseness-1

\textbf{Varying $\boldsymbol{\lambda_P}$ results in disjunct modes} The mean performance differs strongly between the pre-training tasks $\lambda_P$, showing two clear modes as hypothesized in \Cref{sec:zoo_generation}. 
These differences between pre-training tasks carry over to the fine-tuned models (Table \ref{tab:diversity-refined}). Supervised pre-training yields higher performance and a lower generalization gap. These models show correspondingly higher agreement. The performance spread within the pre-training task groups is however relatively close to the overall spread, which shows that models within the two groups do not vary more or less than the overall models. 
As expected, these distinct modes are also recognizable in the model weights directly. Skewness, kurtosis and $L^2$ norm show stark differences between pre-training tasks, both before (Table \ref{tab:diversity-pt}) as well as after fine-tuning (Table \ref{tab:diversity-refined}). Further, each pre-trained model forms a cluster with its fine-tuned derivatives. Pair-wise distances between pre-trained and fine-tuned derivatives are one order of magnitude smaller than the distance between pre-trained models. Lastly, supervised pre-training appears to better align models with fine-tuning, since the distance within supervised clusters is significantly smaller than in contrastive clusters.\looseness-1

\textbf{Varying $\boldsymbol{\lambda_F}$ causes additional clusters and smooth variations}
When considering fine-tuning hyperparameters $\lambda_F$, the optimizer choice has a strong impact on performance, favoring AdamW over SGD. Here, the two modes in performance are even clearer. Interestingly, not only the performance but also the spread of SGD-trained models is smaller, indicating a more homogeneous behavior of these models. The modes are likewise distinct in the weights kurtosis and $L^2$ norm.
The learning rate is expected to create more smooth variation between models. In our experiments, it has little impact on performance, but a surprisingly strong impact on generalization and agreement. Notably, a lower learning rate tends to decrease the generalization gap with lower spread. Contrary to the expectation of higher noise, a higher learning rate leads to a lower accuracy spread. Similar to performance, the variation in weights is continuous and smooth within the different modes, matching expectations.

We note that variation in the generating factors for both pre-training and fine-tuning stages introduce meaningful behavioral diversity in different metrics, even among the well-trained models. As intended, the zoo contains both distinct modes as well as smooth variation both in model behavior as well as in model weights.


\section{Experiments}
\label{sec:applications}

Previous work has shown multiple potential applications for structured populations of neural networks, and has demonstrated they could be leveraged to improve model performance on specific tasks. In this Section, we showcase an exploratory analysis of two novel applications for our model zoo which focus on the two-stage training procedure: model lineage prediction and model weights averaging. Our initial experiments use methods from the literature on our zoos, and show where they perform best and where they break. In \Cref{sec:intended_uses} thereafter, we additionally present possible applications that are not dependent on this two-stage training procedure, and that have been extensively explored on earlier model zoos in the existing literature. By doing this, we demonstrate that our model zoo is both a tool and a benchmark for further improving these methods.

\subsection{Lineage prediction}
\label{sec:lineage}

Fine-tuning from pre-trained models is a common strategy in machine learning. Often, the pre-trained models are provided by a third party and may be taken from a model hub. This creates several challenges, for example around intellectual property, liability, or model certification. For that reason, recent work has shown the importance of better understanding the lineage relation between models in the real world~\citep{horwitzOriginLlamasModel2024,yuNeuralLineage2024}. 
These publications suggest methods to reconstruct the model tree from a set of models, that is, predict whether or not one model is fine-tuned from another model. We call the original model the ``ancestor'', and the one fine-tuned from it the ``descendant''.

\paragraph{Methods} Model Tree Heritage Recover (MoTHeR) is a model lineage prediction method that relies on simple and easily-computed weight features, and that shows very high performance on a model zoo specifically generated for that task~\citep{horwitzOriginLlamasModel2024}.
An alternative method is Neural Lineage, which uses data and linear relaxations of models to approximate a descendant of a model, and subsequently applies behavioral similarity metrics to identify parent-child relations based on the approximations~\citep{yuNeuralLineage2024}. The authors evaluate Neural Lineage on different sets of small MLP and ResNet models with relatively stark variations. In real-world applications, pre-training and fine-tuning are often based on a collection of very few models and recipes, and vary only in minor details. Recognizing such nuances and still identifying relations correctly is therefore of practical importance.
To foster research in that direction, we present our dataset as a challenging testbed for model lineage methods. In the following, we evaluate the MoTHeR approach empirically on our model zoo and evaluate its performance in diverse setups. We describe the method in more detail in Appendix~\ref{ap:mother}.

\begin{wraptable}{r}{0.5\columnwidth}
\vspace{-12pt}
\centering
\small
\setlength{\tabcolsep}{3pt}
\caption{Accuracy and F1 score of model lineage identification using MoTHeR~\citep{horwitzOriginLlamasModel2024} when introducing different variations between the parent and its child models.}
\label{tab:mother}

\begin{tabularx}{0.5\textwidth}{cccccc}
\toprule
& {Epochs} &  {Seed} &  {LR} & {Cls. Head} &  {Optimizer} \\ 
\cmidrule(rl){1-1} \cmidrule(rl){2-2} \cmidrule(rl){3-3} \cmidrule(rl){4-4} \cmidrule(rl){5-5} \cmidrule(rl){6-6}
Acc.                        & 0.9974                                          & 0.9871                                & 0.9870                               & 0.9869                                              & 0.9858                                                   \\
F1                              & 0.8167                                          & 0.0917                                & 0.0833                               & 0.0750                                              & 0.0000                                                   \\ \bottomrule
\end{tabularx}
\end{wraptable}

We evaluate MoTHeR on our ViT model zoo in a set of experiments with increasing complexity. We subsample our ViT zoo to sets of 10 pre-trained models with 12 children for each experiment. In the experiments, we introduce different variations between the children, starting with using different fine-tuning epochs (where the descendants are checkpoints at different fine-tuning epochs, starting from the pre-trained backbone), then varying the random seed, the learning rate (LR), the architecture $\mathcal{A}_F$ of the classification head, and the optimizer. If less than 12 variations are available, we vary epochs to amount to 12 children per pre-trained model. We use MoTHeR to reconstruct the directed relation between the set of models and report the accuracy and F1-score on the adjacency matrix in Table \ref{tab:mother}.\looseness-1

\paragraph{Lineage prediction breaks on small variations in $\boldsymbol{\lambda_F}$}

In the first experiment, the descendants are $12$ checkpoints at different epochs, with all other hyperparameters remaining constant. This yields a set of 10 trees without branches; in other words, we expect every model to be identified as a descendant of the model with the next-lower number of training epochs. In this experiment, MoTHeR achieves high accuracy and a high F1-score of 81.7\%, which indicates that the cluster of pre-trained models are indeed distinct and training trajectories identifiable.

However, when different variations between the children are introduced, e.g., branching on seed, learning rate, classification head, or optimizer, the F1-score drops significantly all the way to 0. The accuracy remains high since there is a large number of true negatives in a sparse adjacency matrix. 
These results indicate that subtle changes between children are difficult to pick up on, and may be confusing for tree identification methods. 
%

\subsection{Weights averaging}
\label{sec:model_weights_averaging}

A more recent application to populations of neural networks comes from the model weights averaging literature. Research has shown that combining multiple neural networks by directly averaging their weights can make them more robust and improve their generalization power~\citep{izmailovAveragingWeightsLeads2019a, guo2023stochastic}. Multiple methodologies exist to perform this combination: averaging different training epochs of a same model~\citep{wortsmanRobustFinetuningZeroshot2022}, or averaging fine-tuned models that originate from the same backbone~\citep{wortsmanModelSoupsAveraging2022, rameModelRatatouilleRecycling2023}. There is also increased interest in re-aligning models to perform weight averaging~\citep{ainsworthGitReBasinMerging2022}.\looseness-1

All of these methods show strong potential, but they perform inconsistently, depending on the experimental setup, the choice of the hyperparameters or the dataset used~\citep{ainsworthGitReBasinMerging2022}. There is therefore a need to explore the conditions under which they succeed. Existing model zoos, because they only include smaller models all trained from scratch, do not allow us to conduct this kind of study.

\begin{figure}[htp]

\centering
    \begin{subfigure}[b]{0.45\textwidth}
        \includegraphics[width=\textwidth]{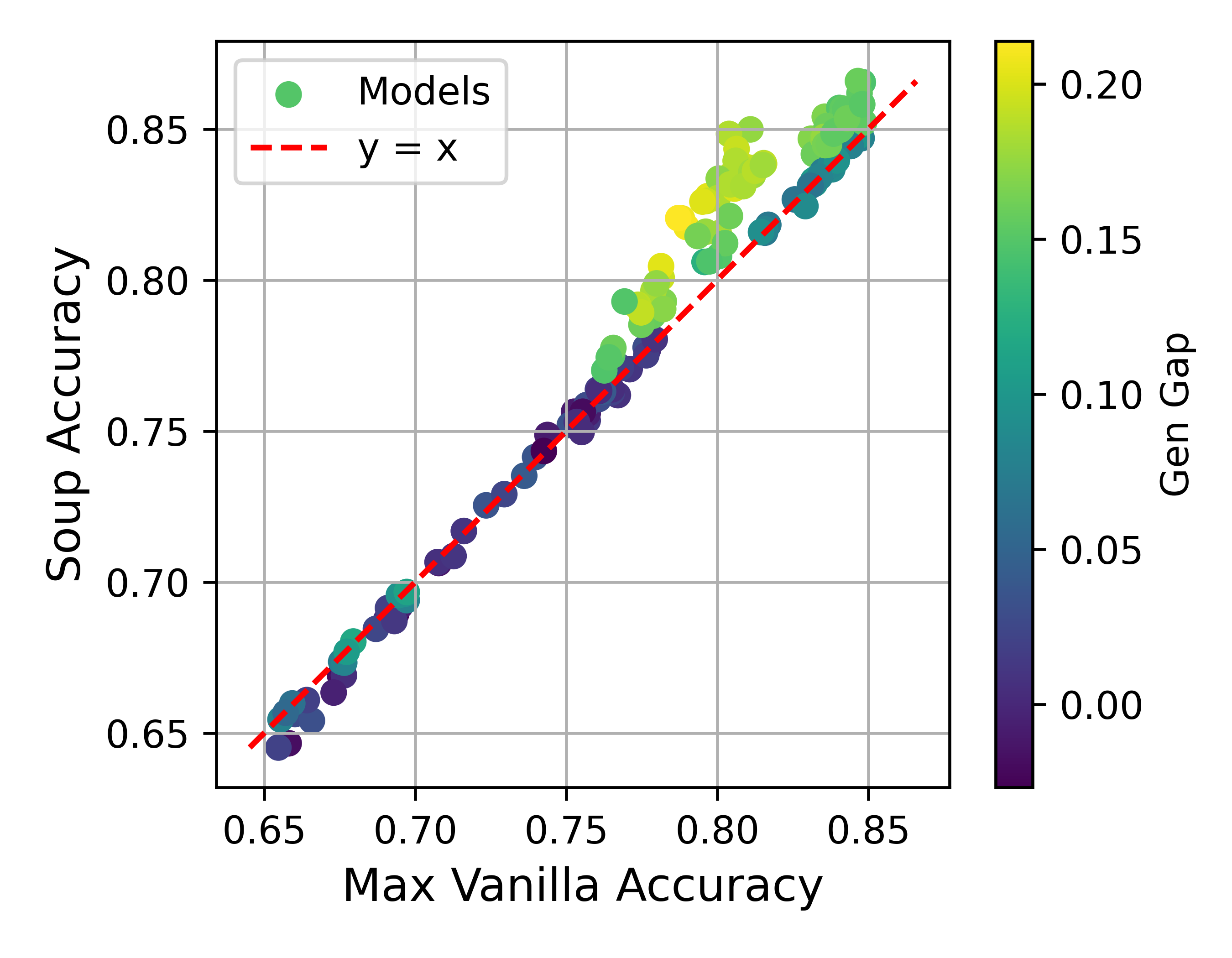}
        \caption{Test accuracy of weight-averaged (last 5 epochs of training) ViTs on CIFAR-100 over max invididual test accuracy.}
        \label{fig:vit_wiseft_max}
    \end{subfigure}
    \hfill
    \begin{subfigure}[b]{0.45\textwidth}
        \includegraphics[width=\textwidth]{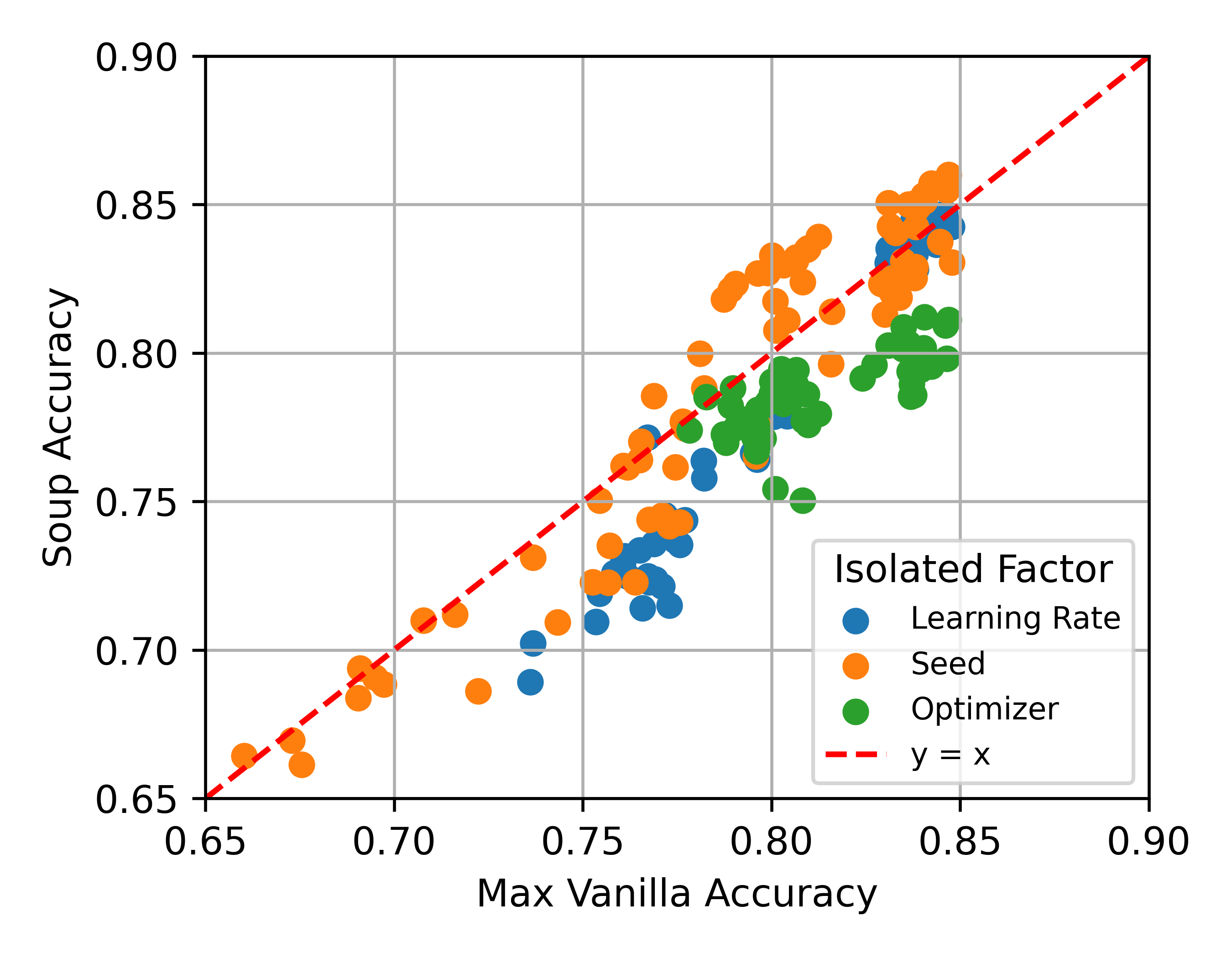}
        \caption{Model soup accuracy over max individual model accuracy. Soups are created by isolating a single hyperparameter variation to average over.}
        \label{fig:vit_soup}
    \end{subfigure}
    \hfill
    \vspace{-0.2cm}
    \caption{Comparison of different weight averaging methods. Averaging over epochs generally increases performance, especially with increasing generalization gap. Averaging over multiple fine-tuned models that share all but one hyperparameter to vary leads to mixed results. Averaging over fine-tuning seeds can improve performance in some cases whereas the other parameters in almost all cases harm performance.}
\end{figure}

As it contains model checkpoints across different epochs, but also multiple fine-tuned models that are based on the same or different backbones, our model zoo fills this gap and allows a comprehensive study of the different existing averaging strategies. The diverse set of hyperparameters in the different zoos will additionally allow researchers to link successful weight averaging to specific sets of generating factors. In the following, we demonstrate the capabilities of this dataset by conducting a preliminary, exploratory study of model weight averaging within the model zoo. 
We consider three averaging dimensions: averaging weights over fine-tuning epochs, averaging over fine-tuning hyperparameters, and averaging over pre-trained models.

\paragraph{Averaging models over fine-tuning epochs}
Averaging models over several stages of training has been shown to improve performance and generalization~\citep{izmailovAveragingWeightsLeads2019a,wortsmanRobustFinetuningZeroshot2022,rameWARMBenefitsWeight2024}. 
To test the benefits of averaging over training trajectories, we evaluate uniform model averaging over the last five epochs of fine-tuning in Figure~\ref{fig:vit_wiseft_max}. This averaging approach yields small but consistent improvements in most cases, even over the maximum performance of 5 averaged epochs. Models that show lower performance before averaging can in some cases decrease slightly in performance whereas models with higher accuracy show better results.
There appears to be a relation between performance gain and generalization gap of the averaged models, which may indicate that averaging reduces overfitting at the benefit of validation performance.
The relation to learning rate mentioned by \cite{wortsmanRobustFinetuningZeroshot2022} is not visible as clearly in our experiments.
Longitudinal averaging results in \Cref{fig:vit_wiseft_over_epochs} in the Appendix show decreased performance toward the beginning of fine-tuning, but show improved performance over individual models after a few training epochs.
\paragraph{Model soups}
Another axis of model averaging is varying fine-tuning hyperparameters, such as learning rate, optimizer or seed. Such combinations have been dubbed \textit{model soups}~\citep{wortsmanModelSoupsAveraging2022}. 
We evaluate such approaches by varying fine-tuning hyperparameters individually: we uniformly average the weights of models that share all but one hyperparameter, e.g., the fine-tuning learning rate. Results in \Cref{fig:vit_soup} show that some soups gain considerable performance while others drop. Averaging over the fine-tuning seed appears to have the highest potential for successful souping but does not always work. Future research may identify conditions for performance improvements more closely.\looseness-1
\paragraph{Git Re-Basin}
The previous two model averaging methods condition the averaged models on a common pre-trained model. Recent work proposed methods to extend these approaches to average models more broadly, by aligning the models to map them to the same subspace in weight space. We evaluate Git Re-Basin --- the alignment method proposed by \cite{ainsworthGitReBasinMerging2022} --- to merge models fine-tuned from different pre-trained models. 
Interestingly, that approach appears to mostly destroy the averaged ViT models. Their performance is around random guessing, no fine-tuning configuration achieved more than 1\% accuracy when averaged over all pretrained models. There are several potential explanations for that behavior. The alignment may not work as well on transformer architectures. Further, averaging several models instead of just two may be too difficult. Lastly, there may be a high loss barrier at the midpoint of the interpolation, but lower-loss regions may still lie on the interpolation lines. In that case, non-uniform averaging may yield better results.\looseness-1
Our findings across different weights averaging methods show large variations in performance. They indicate interesting signals for what type of averaging with what variation improves performance. The lack of success averaging models over pre-trained seeds calls for further research to align and average transformer models. The results show that the model zoo contains relevant variations and challenges to facilitate future research on model averaging.


\section{Intended uses}
\label{sec:intended_uses}

This paper introduces a vision transformer model zoo, designed to support research on neural network populations, particularly focusing on the transformer architecture. It includes multiple pre-trained models with different hyperparameters, and further derives these into several fine-tuned models; this allows the study of both pre-training and transfer learning steps. For both the pre-trained and the fine-tuned models, we provide checkpoints for multiple epochs, making it possible to study the training process. While in \Cref{sec:applications} we showcased possible applications for the zoo made directly possible by the two-steps training scheme --- model lineage prediction and model weights averaging ---, the existing scientific literature suggests multiple additional applications for model zoos in general.

\textbf{Model analysis} The inclusion of model checkpoints opens up opportunities to go beyond simply evaluating models on a test set. Using the CKA similarity~\citep{kornblithSimilarityNeuralNetwork2019} can help compare the representation learned by the different models in the zoos, while leveraging model explanation and visualization is commonly used to better understand single predictions~\citep{karpathyVisualizingUnderstandingRecurrent2015, yosinskiUnderstandingNeuralNetworks2015, bach2015pixel, zintgrafVisualizingDeepNeural2017}. Another strand of research leverages weight statistics to study populations of neural networks and the quality of trained models without requiring a test set~\citep{unterthinerPredictingNeuralNetwork2020, martinPredictingTrendsQuality2021}. On previous zoos, weight statistics performed very well to accurately predict model properties, like performance or hyperparameters~\citep{unterthinerPredictingNeuralNetwork2020, eilertsenClassifyingClassifierDissecting2020, martinPredictingTrendsQuality2021, schurholtHyperRepresentationsGenerativeModels2022, kahana2024deep, schurholt2024SANE}. Some other approaches leverage populations of neural networks to study learning dynamics: they uncover different phases in loss landscapes and show they are good indicators to guide the training procedure~\citep{yangTaxonomizingLocalGlobal2021, yao2018hessian, krishnapriyan2021characterizing, zhou2024MD}.

\textbf{Weight space learning} There is a growing interest in using meta-models to directly learn from neural networks. Early research has shown that simple weight statistics are good predictors for model hyperparameters~\citep{eilertsenClassifyingClassifierDissecting2020} and model performance~\citep{unterthinerPredictingNeuralNetwork2020}. More advanced techniques have since been developed, in particular algorithms that take into account symmetries in weight space~\citep{navonEquivariantArchitecturesLearning2023}, such as Universal Neural Functionals~\citep{zhou2024universal} or Graph-Neural-Networks-based approaches~\citep{kofinas2024graph, lim2023graph}. Such approaches are useful, e.g., for inference attacks on neural networks~\citep{ganju2018property}, or for identifying backdoors in such models~\citep{langoscoDetectingBackdoorsMetaModels2023}.
Similarly, different researchers have developed methods to learn from neural networks in a self-supervised way, allowing the computation of low-dimensional neural network ``hyper-representations''~\citep{schurholtHyperRepresentationsPreTrainingTransfer2022, schurholt2024SANE,
meynentstructure}. All of these approaches leverage populations of neural networks to train their meta-models, and could benefit from a zoo of transformers to extend their methodology to larger models.\looseness-1

\textbf{Model weights generation} Model zoos also provide an opportunity to learn how to directly generate weights for neural networks. While hyper-networks~\citep{haHyperNetworks2016, ratzlaff2019hypergan} do not directly learn from a population of models, some more recent alternatives do leverage populations of already trained neural networks~\citep{knyazevParameterPredictionUnseen2021}. For example, there are works exploring the use of e.g. diffusion models to generate model weights~\citep{peeblesLearningLearnGenerative2022,wang2024neuralnetworkparameterdiffusion,soro2024diffusionbasedneuralnetworkweights, putterman2024learninglorasglequivariantprocessing} 
The self-supervised representation learning approaches represent an alternative to these, as they can be used for both discriminative and generative downstream tasks~\citep{schurholtHyperRepresentationsGenerativeModels2022}. For all these cases, this dataset opens up new opportunities to generate weights for transformer models.

\textbf{Limitations} Conversely, our dataset is not directly aimed at finding best-performing models for computer vision. While the architectures we use are state-of-the-art, the models in our zoo are optimized for diversity, not for performance, as discussed in \Cref{sec:performance}. In addition, while our model zoo may be related to Neural Architecture Search (NAS) datasets such as NAS-Bench~\citep{nasbench101, nasbench201}, its objectives are different. These benchmarks form an excellent basis for the evaluation of NAS methods thanks to the span of architectures and configurations they include, in comparison to which our zoo pales. Nevertheless, our zoo comes with additional data, as it not only includes several performance metrics, but also model weights throughout the training procedure. It can thus be leveraged for a variety of weight-space methods that may not replace traditional NAS approaches, but complement them.
\vspace{-2pt}

\section{Conclusion}
\label{sec:conclusion}
\vspace{-2pt}
Our paper presents the first model zoo of transformer models for computer vision. While previous model zoos primarily consisted of smaller CNNs or a limited selection of ResNet models all trained from scratch, our paper is the first to present a structured population of models based on state-of-the-art architectures, covering both the pre-training and fine-tuning steps. We generated it not necessarily to maximize performance, but rather to provide a large set of diverse models; we validate this diversity through different structural and behavioral metrics. The existing scientific literature contains numerous examples of applications for such populations, and our model zoo will allow them to be more representative of the current advancement of deep learning research. Besides, the two-stage training procedure opens up novel applications, such as model lineage prediction and model weights averaging. We highlight this by proposing an exploratory analysis of these applications, showcasing promising initial results and suggesting a strong potential for improvement, which we reserve for future research. For these reasons, we believe the model zoo presented in this paper to be an essential tool for scaling up population-based methods to real-world applications.

\newpage
\bibliography{bib_manual, bib_auto}
\bibliographystyle{iclr2025_conference}

\appendix
\section{Appendix}

In the Appendix we provide additional information on the model zoo training and performance in the first section. In ~\cref{ap:mother} we describe the MoThHeR method in detail. In ~\cref{sec:dataset_desc} we provide more information about the dataset format. Code and links to the dataset are available on \texttt{\href{https://github.com/ModelZoos/ViTModelZoo}{github.com/ModelZoos/ViTModelZoo}}.

\subsection{Overview of generating factors}
\label{sec:generatingfactors_vit}

\paragraph{Training} During pre-training, the batch size was chosen to accommodate computational limitations, varying between supervised and contrastive models; details are included in Appendix~\ref{sec:pretraining-details}. During fine-tuning, the batch size was set constant to $256$, and when using the SGD optimizer, we use a momentum of $0.99$ in all cases. While the original ViT pre-training spans 300 epochs \citep{dosovitskiyImageWorth16x162020}, we opt for a reduced 90-epoch pre-training as suggested by \cite{beyer2022better}. Differing from \cite{beyer2022better}, we keep the ViT architecture unmodified, to retain broad relevance and consistency.
During pre-training and fine-tuning, we apply consistent data augmentations. We normalize with ImageNet parameters throughout. Following state of the art~\citep{zou2023benefits}, we use a combination of random-cropping to $224\times 224$, MixUp~\citep{Zhang2017}, CutMix~\citep{cutmix} and Random Erasing \citep{Zhong_Zheng_Kang_Li_Yang_2020} for our supervised models to improve generalization. For fine-tuning, we additionally randomly rotate and flip horizontally.

\textbf{Learning Rate}: In order to introduce a controlled and smooth amount of diversity, we fine-tune each model with a set of three different learning rates recommended by previous research on ViT training, $3e-3$ \citep{touvron2021vit-s, steinerHowTrainYour2021, dosovitskiyImageWorth16x162020}, $1e-3$ \citep{steinerHowTrainYour2021, dosovitskiyImageWorth16x162020, huynh2022vit-tiny, beyer2022better} and $1e-4$ \citep{touvron2021vit-s}. For greater simplicity and consistency, we refrain from using learning rate schedulers.

\textbf{Optimizer}: We fine-tune all models using two optimizers, aiming to create distinct modes within clusters of fine-tuned models. As indicated by ~\cite{dosovitskiyImageWorth16x162020}, the Adam optimizer works particularly well with ViTs, and we use its improved version, AdamW \citep{loshchilov2018adamw}. In parallel, we also experiment with SGD, motivated by findings that suggest notable differences in convergence and stability between the two \citep{huynh2022vit-tiny}.

\textbf{Classification Head}: To represent the diversity of approaches in the existing literature, we also vary $\mathcal{A}_F$ and explore the influence of the choice of the classification head on the structure and behavior of the underlying pre-trained model. We fine-tune all pre-trained models with either a linear projection head or a MLP architecture with the hidden layer consisting of 256 neurons and a ReLu activation, followed by the final linear projection layer. 

\textbf{Initialization Seed}: As a control factor we use two different fixed seeds for the classification head initialization. 

Table \ref{tab:generatingfactors_vit} summarizes the grid used for pre-training and fine-tuning. In Table \ref{tab:training-params}, we provide additional details on constant training hyperparameters.

\begin{table}[h]
\centering
\footnotesize
\caption{\textbf{Overview of the generating factors of the model zoo:} We pre-train 10 models based on two configurations (supervised and self-supervised). Each pre-trained model is fine-tuned in 12 configurations and 2 fixed seeds for the classification head, resulting in 240 fine-tuned models in the zoo. Several values for each parameter define a grid within the zoo.}
\label{tab:generatingfactors_vit}
\setlength{\tabcolsep}{4pt}
\renewcommand{\arraystretch}{1.2} 
\begin{tabular}{@{}lcccccc@{}}
& \multicolumn{2}{c}{\textbf{Pre-training}} & \multicolumn{4}{c}{\textbf{Fine-tuning}} \\
\cmidrule(r){2-3} \cmidrule(l){4-7}

\textbf{Architecture $\mathcal{A}$} & \multicolumn{2}{c}{ViT-S 16/224} & \multicolumn{2}{c}{LP head} & \multicolumn{2}{c}{MLP head} \\
\cmidrule(r){2-3} \cmidrule(l){4-7}
\textbf{Dataset $\mathcal{D}$} & \multicolumn{2}{c}{ImageNet-1k} & \multicolumn{4}{c}{CIFAR-100} \\
\cmidrule(r){2-3} \cmidrule(l){4-7}
\textbf{Hyperparameters $\lambda$} & & & & & & \\
\phantom{a} Task & \multicolumn{2}{c}{Supervised, Contrastive} & & \multicolumn{2}{c}{Supervised} & \\
\phantom{a} Learning Rates & \multicolumn{2}{c}{1E-3} & & \multicolumn{2}{c}{3E-3, 1E-3, 1E-4 } & \\
\phantom{a} Weight Decay & \multicolumn{2}{c}{1E-4} & & \multicolumn{2}{c}{0} & \\
\phantom{a} Optimiser & \multicolumn{2}{c}{AdamW} & &\multicolumn{2}{c}{AdamW, SGD} & \\
\phantom{a} Epochs & \multicolumn{2}{c}{90} & & \multicolumn{2}{c}{50} & \\

\end{tabular}
\end{table}

\begin{table}[h]
    \caption{Training hyperparameters for models on ImageNet-1K and CIFAR-100 datasets.}
    \label{tab:training-params}
    \centering
    \begin{tabular}{c c}
        \textbf{Parameter} & \textbf{Value} \\
        \hline
        Stochastic Depth & 0.1 \\
        Label Smoothing & 0.1 \\
        CutMix & 1.0 \\
        MixUp & 0.8 \\
        Random Erase & 0.25 \\
    \end{tabular}

\end{table}

\FloatBarrier

\subsection{Performance metrics for the raw vision model zoo including all models}
\label{sec:ft-acc-overview}

\begin{table*}[h]
    \centering
    \caption{Performance metrics for the raw zoo per pre-training mode averaging the results of different weight initialization seeds and over the different pre-trained models. All metrics are reported in percent.}
    \label{tab:pt-acc-ft}
    \resizebox{\textwidth}{!}{%
    \begin{tabular}{c c c c c c c c c }
        \\
        \textbf{Pre-training} & \textbf{Optimizer} & \textbf{Classification Head} & \textbf{LR} & \textbf{Min} & \textbf{Max} & $\boldsymbol{\mu}$ ($\boldsymbol{\sigma}$) & \textbf{Gen Gap} & \textbf{Agr} \\ 
        \hline
        \makecell{Contrastive \\ Learning} & \makecell{AdamW \\ AdamW \\AdamW \\AdamW \\AdamW \\AdamW \\ SGD \\ SGD\\ SGD\\ SGD\\ SGD\\ SGD} & \makecell{Linear \\ MLP \\ Linear \\ MLP \\ Linear \\ MLP \\ Linear \\ MLP \\ Linear \\ MLP \\ Linear \\ MLP} & \makecell{0.003 \\ 0.003 \\ 0.001 \\ 0.001 \\ 0.0001 \\ 0.0001 \\ 0.003 \\ 0.003 \\ 0.001 \\ 0.001 \\ 0.0001 \\ 0.0001} & \makecell{76.71 \\ 76.92 \\ 76.37 \\ 76.26 \\ 64.94 \\ 65.68 \\ 56.1 \\ 56.4 \\ 49.1 \\ 47.54 \\ 15.41 \\ 7.19} & \makecell{80.42 \\ 79.62 \\ 80.12 \\ 79.58 \\ 69.72 \\ 69.69 \\ 60.72 \\ 60.19 \\ 53.79 \\ 52.11 \\ 20.31 \\ 11.05} & \makecell{78.24 (1.46) \\ 77.96 (1.20) \\ 78.25 (1.63) \\ 77.78 (1.46) \\ 67.39 (1.94) \\ 67.74 (1.72) \\ 58.54 (1.94) \\ 58.64 (1.61) \\ 51.8 (1.94) \\ 50.17 (1.95) \\ 17.04 (1.96) \\ 9.1 (1.44)} & \makecell{16.77 \\ 17.67 \\ 14.67 \\ 14.52 \\ 3.23 \\ 2.88 \\ 0.63 \\ 1.16 \\ -0.59 \\ 0.28 \\ 7.78 \\ 7.54} & \makecell{ 79.83 (1.14) \\ 79.28 (1.09) \\ 80.92 (1.73) \\ 80.48 (1.59) \\ 76.29 (3.50) \\ 76.14 (3.48) \\ 71.72 (5.91) \\ 70.74 (4.70) \\ 67.18 (7.15) \\ 66.18 (6.73) \\ 24.99 (11.76) \\ 11.93 (1.84)} \\
        \midrule
        \makecell{Supervised \\ Learning} & \makecell{AdamW \\ AdamW \\AdamW \\AdamW \\AdamW \\AdamW \\ SGD \\ SGD\\ SGD\\ SGD\\ SGD\\ SGD} & \makecell{Linear \\ MLP \\ Linear \\ MLP \\ Linear \\ MLP \\ Linear \\ MLP \\ Linear \\ MLP \\ Linear \\ MLP} & \makecell{0.003 \\ 0.003 \\ 0.001 \\ 0.001 \\ 0.0001 \\ 0.0001 \\ 0.003 \\ 0.003 \\ 0.001 \\ 0.001 \\ 0.0001 \\ 0.0001} & \makecell{77.83 \\ 78.28 \\ 82.41 \\ 82.01 \\ 81.42 \\ 81.37 \\ 73.61 \\ 71.82 \\ 65.99 \\ 54.02 \\ 2.77 \\ 1.04 } & \makecell{80.97 \\ 81.25 \\ 84.69 \\ 84.65 \\ 84.46 \\ 84.78 \\ 77.71 \\ 77.1 \\ 71.61 \\ 67.64 \\ 6.63 \\ 2.59} & \makecell{79.89 (0.87) \\ 79.80 (0.90) \\ 83.73 (0.65) \\ 83.59 (0.67) \\ 83.12 (0.84) \\ 83.28 (0.93) \\ 76.04 (1.26) \\ 75.1 (1.62) \\ 68.9 (1.86) \\ 62.15 (4.56) \\ 5.24 (1.20) \\ 2.59 (0.38)} & \makecell{19.49 \\ 19.04 \\ 16.1 \\ 16.05 \\ 7.95 \\ 6.9 \\ 0.75 \\ -0.28 \\ -2.11 \\ 0.44 \\ 1.1 \\ -0.4} & \makecell{79.29 (0.90) \\ 78.75 (0.86) \\ 84.15 (0.83) \\ 83.94 (0.87) \\ 86.42 (1.50) \\ 86.46 (1.68) \\ 81.06 (2.75) \\ 80.09 (2.82) \\ 74.56 (3.22) \\ 67.71 (5.24) \\ 2.04 (0.63) \\ 4.18 (4.37)} \\
    \end{tabular}
    }%
    
\end{table*}

In Table \ref{tab:pt-acc-ft} we include detailed performance metrics for all configurations including models with a validation accuracy below 65\%. We report the metrics individually for the different pre-training methods, optimizers and learning rates. Therefore each row consists of the average of the two different weight initialization seeds and for each pre-trained model. In general, we see higher performance with the AdamW optimizer compared to SGD. When using the AdamW optimizer we observe better performance with the supervised models across the board. The AdamW optimizer in general also worked better with lower learning rates, whereas SGD only achieves comparable performance with the higher learning rates. The classification head architecture as well as seed for initialization during fine-tuning only exhibit marginal differences in terms of performance. In most cases, we observe slightly higher variance in terms of modal agreement with contrastive pre-training compared to supervised pre-training. 

\FloatBarrier
\subsection{Pre-training performance details}
\label{sec:pretraining-details}

We analyze three pre-trained models using contrastive learning as well as seven pre-trained models using supervised-learning. Two models were trained with the same seed but varying batch size (256 and 512 respectively). Metrics for the supervised configuration are included in Table \ref{tab:pt-acc-sl} and the contrastive models are shown in Table \ref{tab:pt-acc-cl}.

\begin{table}[h]
    \centering
     \caption{Validation accuracy (cross-entropy loss) on ImageNet-1K for supervised models after 90 epochs of training}
    \label{tab:pt-acc-sl}
    
    \resizebox{0.7\columnwidth}{!}{%
    \begin{tabular}{c c c c}
        \\
        \textbf{Seed} & \textbf{Batch Size} & \textbf{Validation Accuracy} & \textbf{Loss (CE)} \\
        \hline
        42 & 256 & 67.46\% & 1.457 \\
        6312 & 256 & 69.58\% & 1.372 \\
        6312 & 512 & 70.99\% & 1.299 \\
        1957 &  512 &  71.42\% & 1.293 \\
        2517 & 512 & 70.11\% & 1.341 \\
        3753 & 512 & 70.05\% & 1.346\\
        7853 & 512 & 69.02\% & 1.4\\
    \end{tabular}
    }
   
\end{table}

Our best performing model achieved a validation accuracy of 71.42\%. Overall, the supervised models exhibit a difference of up to 3.96 percentage points in terms of validation accuracy. Isolating models where only the seed varies (n=5 for batch size 512) the difference decreases to 2.4 percentage points. An increased batch size (256 vs 512) with equal initialization showed a difference of 1.41 percentage points. Notably, the seed seems to have a bigger impact than the batch size, as one model with a lower batch size still outperformed a model with a bigger batch size with a different seed.

\begin{table}[h]
    \centering
     \caption{Validation accuracy (Info-NCE loss, top 1 predictions) for contrastive models after 90 epochs of training.}
    \resizebox{0.7\columnwidth}{!}{%
    \begin{tabular}{c c c c}
        \\
        \textbf{Seed} & \textbf{Batch Size} & \textbf{Validation Accuracy} & \textbf{Loss (Info-NCE)} \\
        \hline
        42 & 96 & 92.85\% & 0.4753 \\
        6312 & 128 & 93.75\% & 0.457 \\
        2517 & 128 & 93.17\% & 0.4837 \\
    \end{tabular}
    }
   
    \label{tab:pt-acc-cl}
\end{table}

The models using contrastive learning for pre-training exhibit a similar behavior. While the accuracy scores are not directly comparable due to a different loss function we see that the seed used for initialization has the same impact on performance, i.e. seeds that lead to higher accuracy in the supervised pre-training also perform better in the self-supervised set-up.

\FloatBarrier
\subsection{Additional results}

\begin{figure}[h]
    \centering
    \includegraphics[width=0.7\columnwidth]{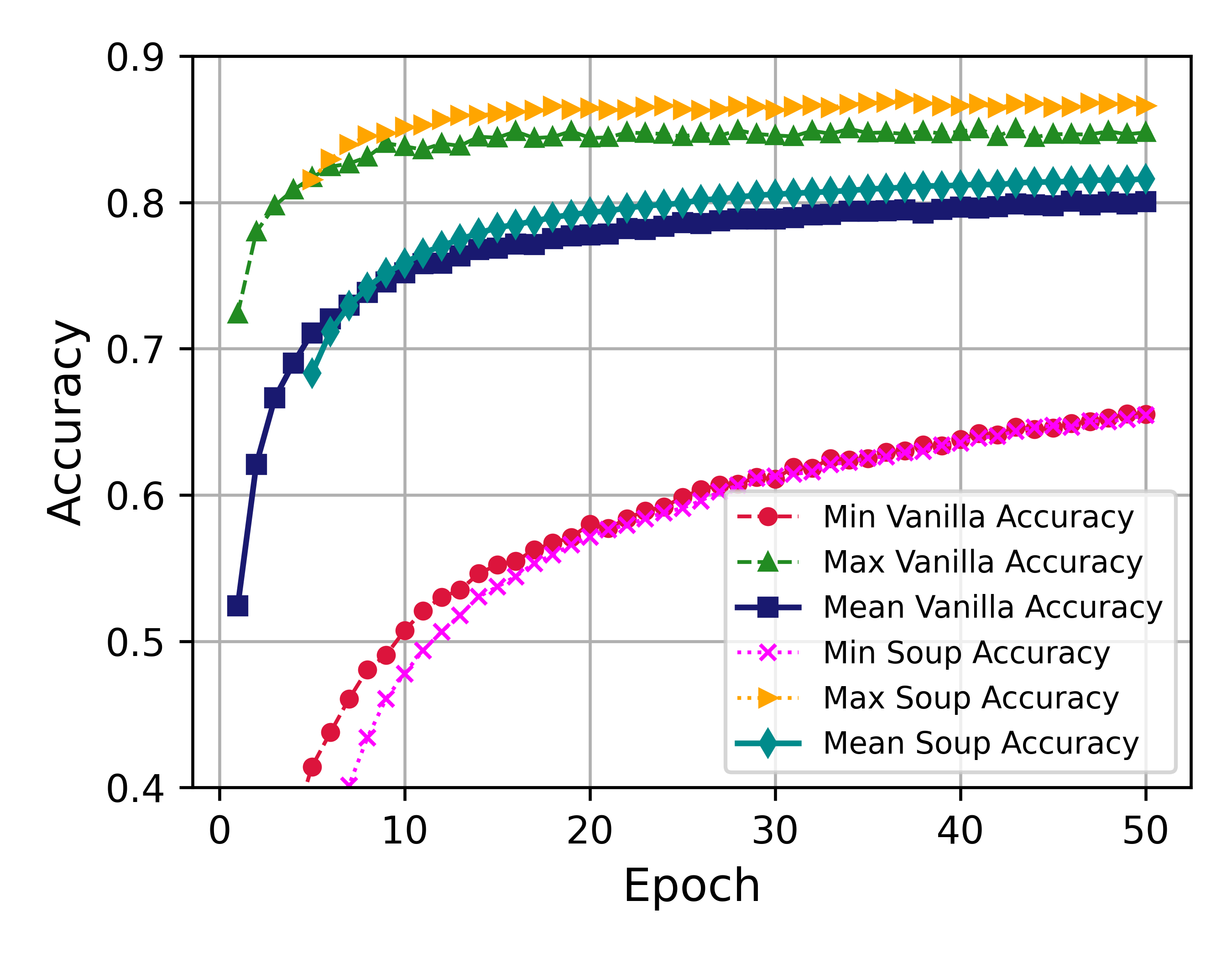}
    \vspace{-0.3cm}
    \caption{Longitudinal test accuracy of weight-averaged ViTs on CIFAR-100 over fine-tuning epochs. Models are averaged over the 5 previous training epochs. Averaging over fine-tuning epochs consistently improves performance after the training curve has bent towards its asymptote.}
    \label{fig:vit_wiseft_over_epochs}
\end{figure}

Figure \ref{fig:vit_wiseft_over_epochs} shows longitudinal weight averaging results compared to the individual model performance. The weights of the checkpoints of the previous 5 epochs of training are averaged starting at epoch 5. The figure shows the minimum, mean, and maximum performance of the weight averaged models in comparison to the vanilla models.

\FloatBarrier

\newpage

\section{MoTHeR method}
\label{ap:mother}

The MoTHeR approach is based on the empirical finding that descendants of the same ancestor form a cluster, and that distinct ancestors form separate clusters, a finding that clearly appears in \Cref{tab:diversity-pt}. As a first step to identify these relations, MoTHeR therefore uses the $L^2$  distance between model weights to cluster models using the $k$-means algorithm~\citep{macqueen1967some}, each corresponding to the same common ancestor. Then, to identify the relation between models within a cluster, MoTHeR proposes to use the kurtosis of the weight matrices. The authors found that kurtosis increases during pre-training, and decreases during fine-tuning. They therefore combine the kurtosis and $L^2$ distance in a new pair-wise distance matrix $\mathbf{M}$, and  use it to build the minimum spanning tree for each cluster. It is built as follows:
\begin{equation}
\mathbf{M} = \mathbf{d}_c + \lambda \cdot \overline{d} \cdot \mathbf{T} + \mathbf{D}_{\infty}
\end{equation}
where \( \mathbf{d}_c \) is the $L^2$ distance matrix for the models in the cluster, with entries \( d_{c,ij} \) representing the pairwise Euclidean distances between the weights of models \( i \) and \( j \) in the cluster. The term \( \lambda \) is a scalar parameter used to weigh the kurtosis-based term, while \( \overline{d} \) is the mean of all the distances in \( \mathbf{d}_c \), given by:
\begin{equation}
\overline{d} = \frac{1}{n^2 - n} \sum_{i \neq j} d_{c,ij}
\end{equation}
where \( n \) is the number of models in the cluster. The matrix \( \mathbf{T} \) is the kurtosis comparison matrix, where each entry \( T_{ij} \) is defined as:
\begin{equation}
T_{ij} = \begin{cases} 
       1, & \text{if } \kappa_i > \kappa_j \\
       0, & \text{otherwise}
   \end{cases}
\end{equation}
where \( \kappa_i \) and \( \kappa_j \) are the kurtosis values for models \( i \) and \( j \), respectively. This matrix identifies pairs of models with increasing kurtosis and adding distance when constructing the tree. Lastly, \( \mathbf{D}_{\infty} \) is a diagonal matrix with infinite values, \( \mathbf{D}_{\infty} = \text{diag}(\infty, \infty, \dots, \infty) \), which ensures that no model can connect to itself in the final tree.

\newpage

\section{Dataset description}
\label{sec:dataset_desc}
\subsection{Generation}

We generate a model zoo for computer vision built on the ViT-S architecture. We train several backbone models with varying hyperparameters, and further fine-tune them using multiple hyperparameter combinations. The details of the dataset generation schemes are described in the main paper, in Section~\ref{sec:zoo_generation}, and the summary of generating factors used can be found in Table \ref{tab:generatingfactors_vit}. Code and links to the dataset are available on \texttt{\href{https://github.com/ModelZoos/ViTModelZoo}{github.com/ModelZoos/ViTModelZoo}}.

We further annotate every model with performance metrics, as detailed in the maintext Section~\ref{sec:performance}. We evaluate on the validation set after every epoch of training and supply results per model in an easily readable JSON format.

We train all models on V100 (16GB) GPUs. In select cases for our supervised pretraining, we use a cluster of 2 GPUs to scale batchsize. In all other cases we use a single GPU. Training times vary per configuration. Pretraining takes rougly 8 days for our contrastive models and 4 days per supervised model.

\subsection{Format}

The dataset is structured as follows:

\dirtree{%
.1 dataset.
.2 pretraining.
.3 model 1.
.4 checkpoint\_0000XY.
.4 config.json.
.4 result.json.
.3 \vdots.
.2 finetuning.
.3 pretrained model 1.
.4 model 1.
.5 checkpoint\_0000XY.
.5 config.json.
.5 result.json.
} 

The pretrained models are stored in the `pretraining` folders, their names include the generating factors used to create them. The finetuned models are stored in the `finetuning` folders and are ordered per pretrained model they were finetuned from. Each model directory contains a `config.json` file with the model configuration as well as a `result.json` file with the model performance metrics. Moreover, it includes checkpoints of the models in the format `checkpoint\_0000XY/checkpoints`.

\subsection{Sensitive Data}

Our NNs are trained on common, publically available datasets. Since our dataset only contains checkpoints of these trained models and performance annotations, it does not contain any personally identifiable information. Similarly, it does not contain any offensive content. Authors bear all responsibility in case of violation of rights.

\subsection{Hosting, availability and license}

Our dataset will be published and made accessible under the Creative Commons Attribution 4.0 International (CC BY 4.0) license.

\end{document}

%% file: math_commands.tex
\usepackage{amsmath,amsfonts,bm}

\def\eqref#1{equation~\ref{#1}}

\def\1{\bm{1}}

\DeclareMathAlphabet{\mathsfit}{\encodingdefault}{\sfdefault}{m}{sl}
\SetMathAlphabet{\mathsfit}{bold}{\encodingdefault}{\sfdefault}{bx}{n}

